\documentclass[conference]{IEEEtran}
\IEEEoverridecommandlockouts
\usepackage{cite}
\usepackage{amsmath,amssymb,amsfonts}
\usepackage{algorithmic}
\usepackage{graphicx}
\usepackage{textcomp}
\usepackage{xcolor}
\usepackage{hyperref}
\usepackage{subfig}
\def\BibTeX{{\rm B\kern-.05em{\sc i\kern-.025em b}\kern-.08em
    T\kern-.1667em\lower.7ex\hbox{E}\kern-.125emX}}
\begin{document}

\title{Hierarchical reinforcement learning for in-hand robotic manipulation using Davenport chained rotations\\
\thanks{This publication has emanated from research supported by Science Foundation Ireland (SFI) under Grant Number SFI/12/RC/2289\_P2, co-funded by the European Regional Development Fund, by Science Foundation Ireland Future Research Leaders Award (17/FRL/4832), and by China Scholarship Council (CSC No.202006540003). For the purpose of Open Access, the author has applied a CC BY public copyright licence to any Author Accepted Manuscript version arising from this submission.}
}

\author{\IEEEauthorblockN{Francisco Roldan Sanchez}
\IEEEauthorblockA{\textit{Dublin City University} \\
\textit{Insight Centre for Data Analytics}\\
Dublin, Ireland \\
francisco.sanchez@insight-centre.org}
\and
\IEEEauthorblockN{Qiang Wang}
\IEEEauthorblockA{\textit{University College Dublin} \\
Dublin, Ireland \\
qiang.wang@ucdconnect.ie}
\and
\IEEEauthorblockN{David Cordova Bulens}
\IEEEauthorblockA{\textit{University College Dublin} \\
Dublin, Ireland \\
david.cordovabulens@ucd.ie}
\and
\IEEEauthorblockN{Kevin McGuinness}
\IEEEauthorblockA{\textit{Dublin City University} \\
\textit{Insight Centre for Data Analytics}\\
Dublin, Ireland \\
kevin.mcguinness@insight-centre.org}
\and
\IEEEauthorblockN{ Stephen J. Redmond}
\IEEEauthorblockA{\textit{University College Dublin} \\
\textit{Insight Centre for Data Analytics}\\
Dublin, Ireland \\
stephen.redmond@ucd.ie}
\and
\IEEEauthorblockN{ Noel E. O'Connor}
\IEEEauthorblockA{\textit{Dublin City University} \\
\textit{Insight Centre for Data Analytics}\\
Dublin, Ireland \\
noel.oconnor@insight-centre.org}
}

\maketitle

\begin{abstract}
End-to-end reinforcement learning techniques are among the most successful methods for robotic manipulation tasks. However, the training time required to find a good policy capable of solving complex tasks is prohibitively large. Therefore, depending on the computing resources available, it might not be feasible to use such techniques. The use of domain knowledge to decompose manipulation tasks into primitive skills, to be performed in sequence, could reduce the overall complexity of the learning problem, and hence reduce the amount of training required to achieve dexterity. In this paper, we propose the use of Davenport chained rotations to decompose complex 3D rotation goals into a concatenation of a smaller set of more simple rotation skills. State-of-the-art reinforcement-learning-based methods can then be trained using less overall simulated experience. We compare this learning approach with the popular Hindsight Experience Replay method, trained in an end-to-end fashion using the same amount of experience in a simulated robotic hand environment. Despite a general decrease in performance of the primitive skills when being sequentially executed, we find that decomposing arbitrary 3D rotations into elementary rotations is beneficial when computing resources are limited, obtaining increases of success rates of approximately 10\% on the most complex 3D rotations with respect to the success rates obtained by a HER-based approach trained in an end-to-end fashion, and increases of success rates between 20\% and 40\% on the most simple rotations. 
\end{abstract}


\section{Introduction}
\label{sec:int}

A primary reason why it takes so long to train state-of-the-art reinforcement learning techniques in manipulation tasks is that they are very complex to solve without domain knowledge \cite{ddpg,mbased,demostration}. However, most manipulation tasks can be decomposed into a number of easier tasks that a robot can learn more easily with much less training experience required \cite{motion_primitives,motion_primitives2}. For example, robots can easily learn how to reach a point in space, or how to push an object, when these tasks are trained independently \cite{prim}. However, if the robot needs to learn from scratch how to both reach for an object and then push it, the training time required increases in a nonlinear manner, meaning it often requires more simulated experience overall to learn these skills \cite{versatile}.

Training time becomes an important matter for tasks that have high degrees of complexity, like the block manipulation tasks in OpenAI's Gym environment \cite{openai}. The most popular method for learning in this kind of goal-based environment is Hindsight Experience Replay (HER) \cite{her} trained in conjunction with Deep Deterministic Policy Gradients (DDPG) \cite{ddpg}. HER is capable of successfully solving most tasks implemented in this environment, but the amount of simulated experience it requires for training is immense, exceeding  $38\times 10^7$ time-steps in the most complex tasks, and even then it is not able to discover an optimal policy that is able to solve all object rotation goals. Furthermore, 19 CPU cores are required to generate simulated experience in parallel \cite{her_results}.

Traditional robotic manipulation systems usually define a set of non-primitive and primitive tasks that have a hierarchy \cite{hierarchy,hierarchy2,hierarchy3}. Non-primitive actions are a composition of primitive actions performed in a particular order, where a primitive action is one which cannot or should not be further divided into a sequence of simpler actions. Therefore, in order to reproduce a non-primitive action, a robot could sequentially perform a set of primitive actions. We refer to Hierarchical Reinforcement Learning (HRL) as those reinforcement learning methods that decompose a problem into a hierarchy of \textit{sub-problems} such that solving the original problem requires the sequential execution of the solutions of these \textit{sub-problems} as if they were primitive actions. 


In order to demonstrate the benefit of using domain knowledge to decompose a manipulation problem into primitive sub-problems, in this paper we propose the use of Davenport angles \cite{davenport} to decompose 3D rotations of a cube into a sequential chain of 1D rotations. By independently learning to perform 1D rotations around different orthogonal axes, which can then be executed sequentially to achieve a 3D rotational goal, we investigate if this methodology will accelerate the overall reinforcement learning goal when compared to training using the standard learning method that uses HER.

\section{Related work}
\label{sec:rw}

\subsection{Hindsight Experience Replay}



HER is an exploration strategy that allows agents to learn faster, particularly when dealing with sparse rewards (i.e., temporally intermittent rewards). More specifically, HER has proven to be a very efficient method in reward settings where the reward is sparse and does not carry much information, such as binary rewards. 

In a standard reinforcement learning framework, the agent learns from positive experiences \cite{onpolicy}. In other words, under a standard framework, every time that agent actions lead to an unsuccessful example where the task is not correctly solved, the only outcome the agent can have is that the sequence of actions that was taken did not lead to the target goal. However, if the objective would have been to reach the achieved result, this same sequence of actions would then be considered a valid sequence from which the agent could learn.

HER facilitates learning from failed attempts by replaying each episode but altering the goal that the agent was originally trying to solve. For example, if a robotic hand is trying to manipulate a cube so that it reaches a target pose $p$, but the actions taken lead to a different pose $p'$, when HER replays, the agent would get two outcomes: that this sequence of actions is not optimal to reach pose $p$, but instead it is optimal to reach the pose $p'$. In other words, the agent will be learning from mistakes. 


Even though HER is a method that is capable of solving complex manipulation tasks, it takes a very long time to train because of the lack of domain knowledge employed during the learning phase. In this paper we explore this problem in the context of in-hand cube manipulation tasks by decomposing the target goals and training agents (using HER) to learn simpler primitive skills.

\subsection{Hierarchical control}

While HER is a model-free method that allows agents to learn dexterous manipulation skills using minimal domain knowledge, traditional robotic manipulation controllers make use of a simpler set of manipulation abilities (primitive actions) that, when sequentially executed, can solve more complex tasks. 

In the context of in-hand robotic manipulation, these primitives are usually defined depending on the type of action they produce. For example, Li \textit{et al.} define three different skills (reposing, sliding and flipping), and a mid-level controller plans how to sequentially apply them to reach a particular cube pose \cite{hier_control1}. Bhatt \textit{et al.} use a similar method but define those primitives as shift, pivot, twist, and finger gait, determining how the robot fingers behave for each of the primitives \cite{hier_control2}.

Instead of focusing on specific hand movements, in this paper we propose to decompose the target pose of a cube into a sequence of simpler rotations that are easier to learn by an agent using a a model-free method such as HER+DDPG, and sequentially apply these learned actions to successfully complete the more complex overall 3D rotation.

\section{Method}
\label{sec:method}

Complex tasks are often defined by complex goals. In the case of the OpenAI Gym hand block manipulation environment, this complexity appears when the goal the robot needs to achieve consists of implementing arbitrary 3D rotations. 

\begin{figure}[!t]
\centering
\includegraphics[width=0.5\textwidth]{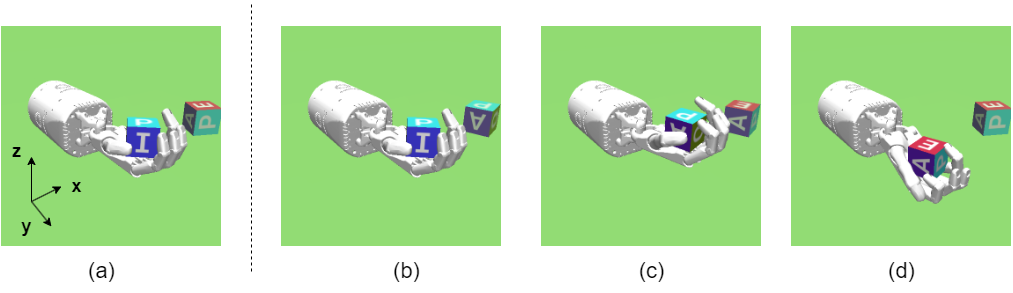}
\caption{Example of a 3D rotation goal decomposition into three Davenport angles. A target 3D rotation (a) is decomposed into three consecutive target rotations around the  (b) $z$, (c) $x$ and (d) $z$ axes, respectively.} 
\label{fig:davenport}
\end{figure}

However, any 3D rotation can be decomposed into three sequential rotations around the orthogonal axes of a known reference frame \cite{euler_angles}. These rotations can be performed around the axes of a fixed system of reference (extrinsic rotations) or around the axes of the rotating system of reference (intrinsic rotations), and only if the second axis of rotation is orthogonal to the plane containing the first and third axes \cite{goldstein}. In the case of the OpenAI's robotic hand environment, this reference frame is fixed, and therefore, the decomposed sequence of rotations applied to the cube must be extrinsic (see Figure \ref{fig:davenport}).

The OpenAI Gym robotic hand block environment contains three different manipulation tasks of increasing complexity:

\begin{itemize}
    \item \textit{RotateZ:} goal consists of a random target rotation around the $z$ axis of the reference frame and position is ignored.
    \item \textit{RotateParallel:} goal consists of a random target rotation around the $z$ axis and rotations by multiples of $\frac{\pi}{2}$ around the $x$ and $y$ axes, with position being ignored. 
    \item \textit{RotateXYZ:} goal consists of a random target 3D rotation with the target position ignored.
\end{itemize}

Observations in all of these tasks include positions and velocities of the robot joints and the pose of the object manipulated (its Cartesian position and its rotation expressed in quaternions).

The reward system applied in this environment is as follows: at each time-step, the agent gets 0 reward if the task is completed, and -1 otherwise. In order to determine whether a rotation has been completed successfully, a quaternion difference between the desired and the achieved poses is used, $\theta =\arccos (2 \langle q_{1}, q_{2}\rangle^2 - 1)$, where $\langle q_{1}, q_{2}\rangle$ represents the inner product of quaternions $q_{1}$ and $q_{2}$, and is considered successful when this difference is less than 0.1 radians.


In order to learn rotations around the $x$ and $y$ axes, two different new tasks based on the \textit{RotateZ} task were implemented for the hand environment\footnote{Tasks can be found at \url{https://github.com/franroldans/custom_gym}}:

\begin{itemize}
    \item \textit{RotateX:} goal consists of a random target rotation around the $x$ axis of reference frame and again position is ignored.
    \item \textit{RotateY:} goal consists of a random target rotation around the $y$ axis of reference frame, with the block's position ignored;
\end{itemize}

\subsection{Policy learning}
We trained HER in combination with DDPG (denoted HER+DDPG) for each of the basic uni-axial rotation tasks: \textit{RotateX}, \textit{RotateY}, and \textit{RotateZ}. For these tasks, episodes have a length of 100 time-steps. This means that, during inference, the agent will have 300 time-steps in order to solve the task, 100 time-steps for each of the three chained rotations.  

To have a fair comparison, we only train HER+DDPG for the \textit{RotateParallel} and \textit{RotateXYZ} tasks using the same number of time-steps used for training the three policies that perform rotations around extrinsic axes, which are the policies trained on the \textit{RotateX}, \textit{RotateY} and \textit{RotateZ} tasks: 4,000,000 + 4,000,000 + 2,000,000 = 10,000,000 time-steps. This way, we can determine whether using Davenport chained rotations is comparable to end-to-end learning of 3D rotations in terms of the accuracy with which the target rotation is achieved for the same amount of training. 

We hypothesize that training HER+DDPG to learn rotations around extrinsic axes and using the policies learned in a sequential manner can achieve better success rates than using HER in an end-to-end fashion on the most complex \textit{RotateParallel} and \textit{RotateXYZ} tasks, if both methods are trained using the same amount of simulated experience.

\subsection{Evaluation}

There is no unique way to decompose a 3D rotation into three consecutive rotations around extrinsic axes. The same rotation can be achieved by decomposing a 3D rotation into elementary rotations around $z{\text -}x{\text -}z$, $x{\text -}y{\text -}x$, $y{\text -}z{\text -}y$, $z{\text -}y{\text -}z$, $x{\text -}z{\text -}x$ or $y{\text -}x{\text -}y$. 


To evaluate our method and compare its performance against that obtained with end-to-end learning, we create a test set of 6,600 different initial and target cube poses for each rotation configuration by random sampling on the goal space. This test set is divided depending on how many extrinsic rotations the robot must perform to achieve the target goals. 

 Not all rotations that are comparable to the \textit{RotateXYZ} task can be performed by the policy trained on the \textit{RotateParallel} task because the goals generated on the \textit{RotateParallel} task only had angles of rotation around the $x$ and $y$ axes contained in the set $\{ -\pi, -\frac{\pi}{2}, 0, \frac{\pi}{2}, \pi \}$.  The distribution of target goals of this test set is shown in Table \ref{testset}.

\begin{table}[!t]
\caption{Test set created to evaluate the method. Table shows how many poses are in the test set that can be solved with either 1, 2 or 3 extrinsic rotations.}
\begin{center}
\setlength{\tabcolsep}{10pt}
\renewcommand{\arraystretch}{1.5}
\begin{tabular}{|c|c|c|c|}
\hline
\textbf{Comparable}&\multicolumn{3}{|c|}{\textbf{\# rotations}} \\
\cline{2-4} 
\textbf{to} & 1  & 2  & 3  \\
\hline 
\textit{RotateParallel}& 200 & 1000 & 2000 \\
\hline 
\textit{RotateXYZ}& 600 & 2000 & 4000 \\
\hline 
\end{tabular}
\label{testset}
\end{center}
\end{table}

\section{Results}
\label{sec:results}

\begin{figure*}[!t]

     \centering
  \subfloat[\label{1a}]{%
       \includegraphics[width=0.25\textwidth]{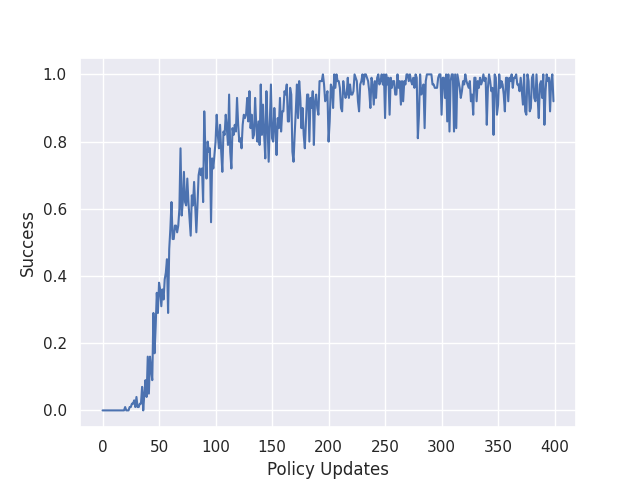}}
    \centering
  \subfloat[\label{1b}]{%
        \includegraphics[width=0.25\textwidth]{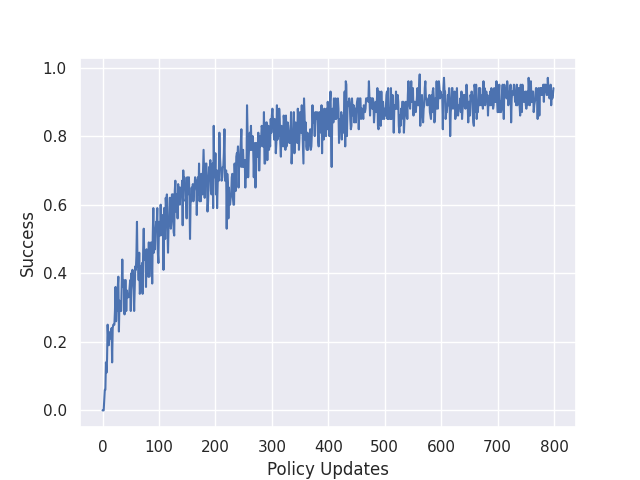}}
     \centering
  \subfloat[\label{1c}]{%
        \includegraphics[width=0.25\textwidth]{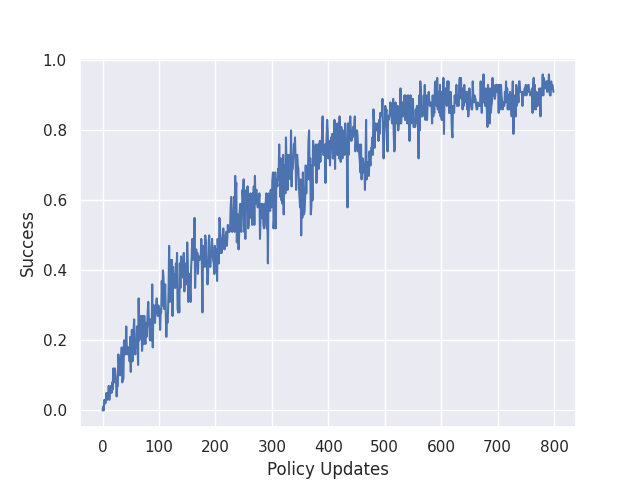}}
    \hfill
        \caption{Success rate evolution during training of the policies trained for the original (a) \textit{RotateZ}, and the implemented (b) \textit{RotateX} and (c) \textit{RotateY} tasks. Shown is success rate versus policy update count, which happens every 50 episodes (5,000 time-steps).} 
        \label{fig:primitives_curves}
\end{figure*}

As predicted, training a policy using HER in conjunction with DDPG was able to successfully learn rotations around extrinsic axes. However, in order to achieve near perfect success rates, the \textit{RotateX} and \textit{RotateY} tasks needed a total of 40 epochs of training instead of the 20 epochs that it takes for the \textit{RotateZ} task (see Figure \ref{fig:primitives_curves}). This happens because rotations around the $z$-axis always have one of the cube faces parallel to the palm of the hand, which makes the task easier, as the cube position is very stable. In contrast, rotations around the $x$ and $y$ axes often have cube poses where there is no face parallel to the palm of the hand, making it more challenging to maintain a stable grasp on the cube.

\begin{figure*}[!t]
     \centering
      \subfloat[\label{1a}]{%
       \includegraphics[width=0.25\linewidth]{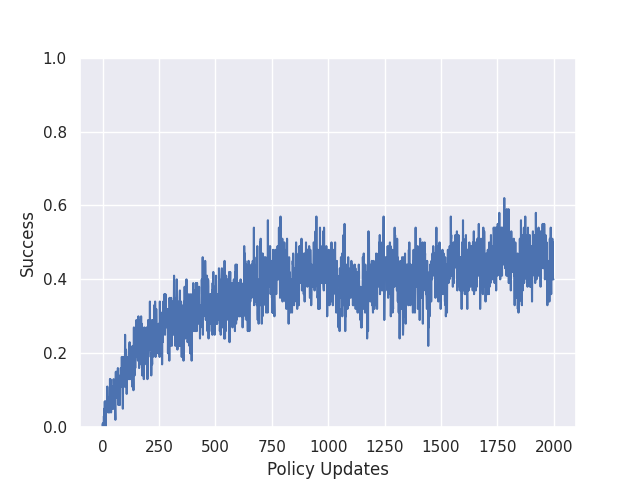}}
    \centering
  \subfloat[\label{1b}]{%
        \includegraphics[width=0.25\linewidth]{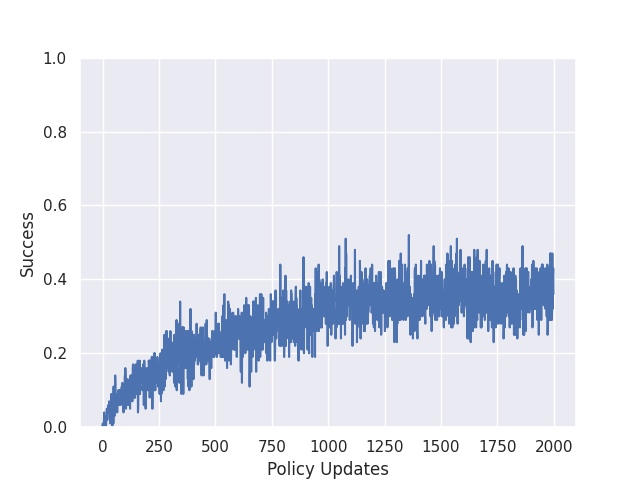}}
    
        \caption{Success rate evolution per policy update of HER trained in an end-to-end manner for the (a) \textit{RotateParallel}, and the (b) \textit{RotateXYZ} tasks using $10^7$ time-steps of simulated experience. }
        \label{fig:non_primitives_curves}
\end{figure*}

These initial experiments suggest that it is indeed the case that chained rotations can achieve better performance than end-to-end training. Using the same overall amount of time-steps for training rotations around intrinsic axes can achieve approximately 100\% success rate, while for the \textit{RotateParallel} and the \textit{RotateXYZ} tasks, end-to-end HER can only achieve approximately 50\% and 40\% of success rate, respectively (see Figure \ref{fig:non_primitives_curves}). 

Results after evaluating the method on the test set show that the hypothesis stated in this paper is supported: decomposing complex 3D rotations into elementary rotations around extrinsic axes can be beneficial when computing resources for training are limited (see Table \ref{scores1} and \ref{scores2}). However, the success rates obtained when applying this approach vary depending on how many elementary extrinsic rotations must be applied to achieve the target 3D rotation. 

\begin{table}[!t]
  \caption{Success rates on the test set for our method and the policy learned on the \textit{RotateXYZ} experiment depending on how many extrinsic rotations the robot is required to perform. }
  \label{scores1}
  \setlength{\tabcolsep}{10pt}
  \centering
  \begin{tabular}{c|ccc}
    Method & 1 rotation & 2 rotations & 3 rotations \\
    \hline 
    End-to-end HER  & 57.5\%  & 47.25\% & 43.85\%   \\ 
    \hline 
    $z-x-z$  & \textbf{96.16\%} & 82.21\% & 51.68\% \\
    $z-y-z$  & \textbf{96.16\%} & \textbf{88.89\%} &\textbf{51.80\%}  \\
    $x-y-x$ & 93\% & 71.13\% & 48.75\% \\
    $x-z-x$ & 93\% & 80.35\% & 49.07\% \\
    $y-x-y$ & 90.50\% & 73.10\% & 45.65\% \\
    $y-z-y$ & 90.50\% & 79.65\% & 49.42 \% \\
  \end{tabular}
\end{table}

\begin{table}[!t]
  \caption{Success rates on the test set for our method and the policy learned on the \textit{RotateParallel} experiment depending on how many extrinsic rotations the robot is required to perform. }
  \label{scores2}
  \setlength{\tabcolsep}{10pt}
  \centering
  \begin{tabular}{c|ccc}
    Method & 1 rotation & 2 rotations & 3 rotations \\
    \hline 
    End-to-end HER  & 69\%  & 63.4\% & 47.15\%   \\ 
    \hline 
    $z-x-z$  & \textbf{97.5\%} & 88.40\% & 66.05\% \\
    $z-y-z$  & \textbf{97.5\%} & \textbf{89.30\%} &\textbf{68.20\%}  \\
    $x-y-x$ & 92\% & 80.10\% & 62.65\% \\
    $x-z-x$ & 92\% & 83.60\% & 62.90\% \\
    $y-x-y$ & 94\% & 82.90\% & 60.65\% \\
    $y-z-y$ & 94\% & 85.20\% & 62.30 \% \\
  \end{tabular}
\end{table}

\section{Discussion}

While the success rate obtained when requiring only one extrinsic rotation is near perfect, this result does not hold when two or more extrinsic rotations are needed. The reason behind this decrease in performance is because each rotation step expects the previous rotation to have been performed perfectly. If the second extrinsic rotation is not achieved, the third rotation will automatically fail because the policy employed has been trained to perform rotations around only one axis, but the goal it needs to reach would also require it to correct the previous mistake. 

There are two main possible alternatives to overcome this problem. First, data augmentation could be used in order to add noise to the initial cube positions so that the primitive policies learned could also take into account errors on the previous steps \cite{data_aug,data_aug2}, helping the robot achieve a better performance at the expense of more training time. Second, the method of decomposing the rotation could be applied in an iterative manner, with each sequence of three rotations converging on the target goal through repeated attempts. However, doing this would obviously slow down the execution of the manipulation task.

From a qualitative analysis of some of our preliminary results, most of the errors our method produced in the second rotation appeared when dealing with large rotations. While the policies learned for the \textit{RotateX} and \textit{RotateY} tasks were very proficient in applying rotations on the range of $[ -\frac{\pi}{2}, \frac{\pi}{2}]$, when having to apply rotations in the range of $[ -\pi, -\frac{\pi}{2})$ and $(\frac{\pi}{2},\pi]$, they were not always correctly executed. This was overcome by introducing further intermediate steps: whenever these policies need to perform a large rotation with a magnitude greater than $\frac{\pi}{2}$, firstly rotate the cube either $\frac{\pi}{2}$ or $-\frac{\pi}{2}$ and then apply the remaining smaller rotation to complete the goal.

After including this extra intermediate step, the success rate on the test set increased from 67.05\% to 88.89\% for those 3D rotations that can be decomposed into two extrinsic rotations in the $z-y-z$ configuration. Some errors appeared in the third extrinsic rotation, where the agent could not always maintain or deal with tilted cube poses. This happens because the policy trained on the primitive tasks had the cube initialized in a way that it always had one face touching the robotic hand in a stable position. Because the policy never saw this kind of tilted configuration during training, on execution it tends to fail. A solution to this would be introducing these tilted initial configurations and goals into the environments so that the robot sees these configurations during the learning phase. Another solution could be forcing the agent to learn how to apply rotation vectors instead of reaching a particular pose, making the learning invariant to the initial and target poses. Another possible solution could be applying intermediate rotations with a magnitude of $\frac{\pi}{2}$, where the cube would be kept face down for all intermediate rotations until the last rotation, and then apply the remaining rotation in the last step. 

As mentioned in Sections \ref{sec:int} and \ref{sec:rw}, HER+DDPG is able to successfully solve this task when being trained for $38\times 10^7$ time-steps and using 19 CPU cores generating simulated experience in parallel, one worker for each core \cite{her_results}. Translated to days, this equates to approximately 30 days of training when using an Intel Core i7-6850K CPU @ 3.60GHz. Considering the sensitivity of reinforcement learning models to hyperparameter selection, this becomes an issue when computing resources are scarce \cite{hyperparameters}. Instead, our method is trained using the same CPU model with no parallelization, and the training of each primitive skill takes less than one day. 

\section{Conclusion}

Training time is a key consideration for all artificial intelligence research, as often small companies and academia have scarce computing resources. In this paper we explored the idea of decomposing complex manipulation tasks into a set of simpler tasks, in the context of a simulated robotic hand environment, where the skills required to complete the simplified tasks can be learned more quickly. Through the addition of domain knowledge, the robot learns how to rotate a cube, with the simulated experience used during training limited to $10^7$ time-steps.

Furthermore, several ideas for future work have been proposed throughout the discussion section above. These are summarized as follows for the reader's convenience:

\begin{itemize}
    \item Use data augmentation techniques to find policies capable of correcting mistakes made in previous intermediate rotation steps;
    \item Iteratively decompose the target rotation into three new elementary extrinsic rotations when an attempt fails;
    \item Improve goal and state exploration by introducing tilted cube configurations and goals during training in order to make the agent learn more unconventional rotations;
    \item Encourage the agent to learn how to apply a given rotation vector instead of reaching a particular pose. 
\end{itemize}



\end{document}